\documentclass[conference]{IEEEtran}
\IEEEoverridecommandlockouts

\usepackage{cite}
\usepackage{amsmath,amssymb,amsfonts}
\usepackage{algorithmic}
\usepackage{graphicx}
\usepackage{textcomp}
\usepackage{xcolor}
\usepackage{multirow}
\usepackage{booktabs}
\usepackage{stfloats}
\graphicspath{{./}}

\def\BibTeX{{\rm B\kern-.05em{\sc i\kern-.025em b}\kern-.08em
    T\kern-.1667em\lower.7ex\hbox{E}\kern-.125emX}}

\begin{document}

\title{DAONet-YOLOv8: An Occlusion-Aware Dual-Attention Network for Tea Leaf Pest and Disease Detection}

%
%
%
%
%

\author{
	Yefeng Wu\textsuperscript{1}\textsuperscript{,}\textsuperscript{2}\textsuperscript{,}\textsuperscript{*}, Shan Wan\textsuperscript{1}\textsuperscript{,}\textsuperscript{2}, Yecheng Zhao\textsuperscript{1}\textsuperscript{,}\textsuperscript{2}, Ling Wu\textsuperscript{1}\textsuperscript{,}\textsuperscript{2} \\
	\textsuperscript{1}\textit{Electronic Science and Technology} \\ 
	\textsuperscript{2}\textit{Anhui University, Hefei, China} \\
	\textsuperscript{*}Emails: wuyefengflc@163.com
}

\maketitle

\begin{abstract}
Accurate detection of tea leaf pests and diseases in real plantations remains challenging due to complex backgrounds, variable illumination, and frequent occlusions among dense branches and leaves. Existing detectors often suffer from missed detections and false positives in such scenarios. To address these issues, we propose DAONet-YOLOv8, an enhanced YOLOv8 variant with three key improvements: (1) a Dual-Attention Fusion Module (DAFM) that combines convolutional local feature extraction with self-attention based global context modeling to focus on subtle lesion regions while suppressing background noise; (2) an occlusion-aware detection head (Detect-OAHead) that learns the relationship between visible and occluded parts to compensate for missing lesion features; and (3) a C2f-DSConv module employing dynamic synthesis convolutions with multiple kernel shapes to better capture irregular lesion boundaries. Experiments on our real-world tea plantation dataset containing six pest and disease categories demonstrate that DAONet-YOLOv8 achieves 92.97\% precision, 92.80\% recall, 97.10\% mAP@50 and 76.90\% mAP@50:95, outperforming the YOLOv8n baseline by 2.34, 4.68, 1.40 and 1.80 percentage points respectively, while reducing parameters by 16.7\%. Comparative experiments further confirm that DAONet-YOLOv8 achieves superior performance over mainstream detection models.
\end{abstract}

\begin{IEEEkeywords}
Tea Leaf Pest, Disease Detection, DAONet-YOLOv8, Object Detection
\end{IEEEkeywords}

\section{Introduction}

Tea is a traditional Chinese specialty and an important economic crop, but pests and diseases seriously threaten yield and quality. Tea trees are vulnerable to anthracnose, red leaf disease, tea sooty mold, tea geometrid pests and white spot disease, which reduce production and degrade quality, causing substantial economic losses. Current control mainly relies on chemical pesticides, and diagnosis still depends on large-scale manual inspection by experienced experts or tea farmers. This process is time-consuming, labor-intensive and highly subjective, making large-scale standardized monitoring and early warning difficult, especially in large plantations where the optimal treatment window is easily missed. Therefore, accurate and timely detection of tea leaf pests and diseases is crucial for improving yield and quality, and intelligent detection technology is urgently needed to support precise pesticide application and green pest management.

With the rapid development of computer vision, image-based automatic detection of tea leaf pests and diseases has achieved notable progress. Early studies mainly relied on traditional machine learning. Hand-crafted features such as color, texture and shape were extracted from images, and then classic classifiers such as random forests were used to recognize pests and diseases. These methods can work reasonably well when features are obvious and the dataset is small, but they have clear limitations. On the one hand, feature design and extraction are heavily dependent on domain expertise, which is costly and inefficient. On the other hand, when facing large variations in sunlight and complex plantation environments, recognition performance degrades significantly; the models have poor robustness and generalization and fail to meet the requirements of practical deployment. 

In recent years, the success of deep learning has made high-precision pest and disease detection possible. Since AlexNet\cite{krizhevsky2012imagenet} won the 2012 ImageNet image classification challenge, deep neural networks have been able to automatically learn discriminative features from raw data, eliminating the need for extensive manual feature engineering and fundamentally overcoming the bottlenecks of traditional machine learning. Building on this, deep learning has also achieved remarkable progress in object detection. Early two-stage detectors such as R-CNN and Faster R-CNN\cite{girshick2014rich,ren2015faster} offer high detection accuracy, but the separate region proposal and classification-regression stages make them too slow for real-time applications. To meet real-time requirements, one-stage detectors have emerged. In particular, YOLO (You Only Look Once)\cite{redmon2016you} formulates object detection as a regression problem and achieves a favorable balance between speed and accuracy. Over years of evolution from YOLOv1 to YOLOv8 released by Ultralytics\cite{redmon2017yolo9000,farhadi2018yolov3,bochkovskiy2020yolov4,jocher2020yolov5,li2022yolov6,wang2022yolov7,ultralytics2023yolov8}, the YOLO family has continuously refined its network architectures and feature fusion strategies, further improving detection accuracy and speed. Recently, CNN- and YOLO-based models have been successfully applied to various crop pest and disease detection tasks.

Despite the success of modern detectors on generic benchmarks and several crops, directly applying off-the-shelf models to tea leaf pest and disease detection remains difficult. Tea plantations exhibit complex backgrounds and unstable illumination, tea pests and diseases are small, visually subtle and often highly similar, and dense branches and leaves frequently occlude key lesion regions. These factors cause generic detectors to suffer from missed and false detections and reduced robustness in real plantations. It is therefore crucial to adapt and enhance deep models to better handle complex backgrounds, perceive tiny lesions and recognize occluded targets while maintaining real-time performance.

\section{Methodology}

\begin{figure*}[!b]
	\centering
	\includegraphics[width=0.9\textwidth]{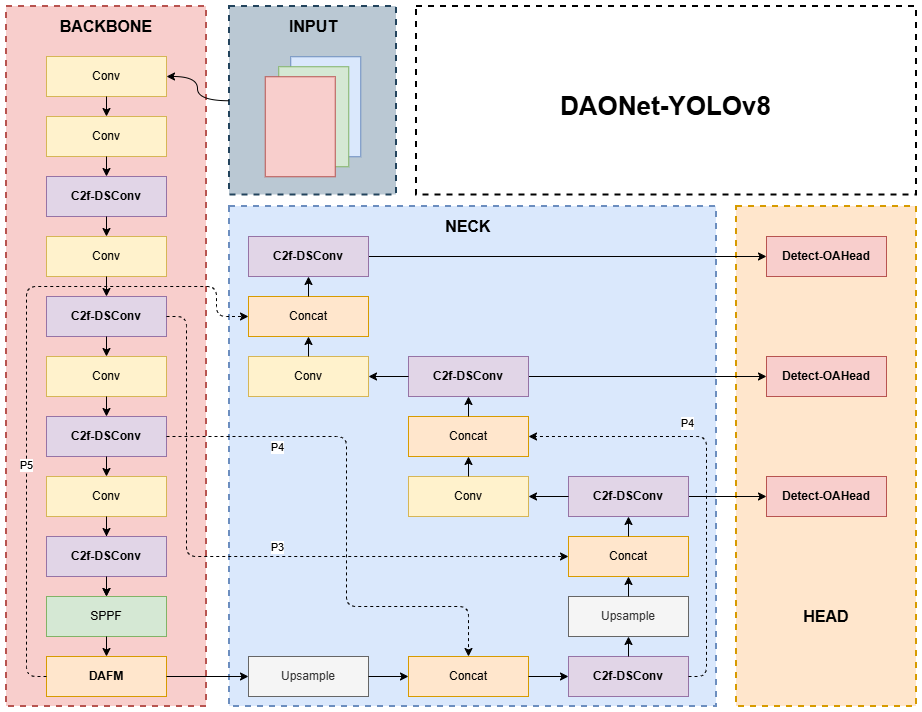}
	\caption{Overall architecture of the DAONet-YOLOv8 network}
	\label{fig:architecture}
\end{figure*}

\subsection{DAONet-YOLOv8}

YOLOv8 consists of a backbone for feature extraction using C2f modules with cross-stage connections, a neck that fuses multi-scale features via FPN and PAN, and an anchor-free decoupled detection head. However, directly applying YOLOv8 to tea leaf pest and disease detection faces challenges: tiny early-stage lesions are easily confused with background clutter, and occlusions among branches yield incomplete features. To address these issues, we propose DAONet-YOLOv8, whose architecture is shown in Fig. 1. It introduces three key components: (1) a Dual-Attention Fusion Module (DAFM) inserted after the SPPF block, combining convolutional local features with self-attention based global context to focus on subtle lesion regions while suppressing background noise; (2) Detect-OAHead, an occlusion-aware detection head that analyzes occlusion patterns and compensates for missing information; and (3) C2f-DSConv blocks that adaptively synthesize convolution kernels of different shapes to capture irregular lesion boundaries. Together, these modules enable DAONet-YOLOv8 to maintain high accuracy and efficiency in complex tea plantation scenes.

\subsection{DAFM: Dual-Attention Fusion Module}
In natural tea plantation environments, images of pests and diseases exhibit highly complex and variable backgrounds, accompanied by lighting changes and background noise. Many types of tea leaf pests and diseases share similar sizes, shapes and textures, and some early-stage lesions are small and inconspicuous. These factors pose stringent requirements on the model''s attention mechanism. The SPPF module in YOLOv8 fuses features from different receptive fields via multi-scale pooling and thus improves multi-scale perception. However, SPPF treats all spatial locations and channels uniformly, without explicitly emphasizing informative features. As a result, when dealing with complex backgrounds, the model may spread attention across irrelevant regions and overlook tiny yet critical lesion features, thereby degrading detection accuracy.

\begin{figure}[htbp]
	\centering
	\includegraphics[width=\columnwidth]{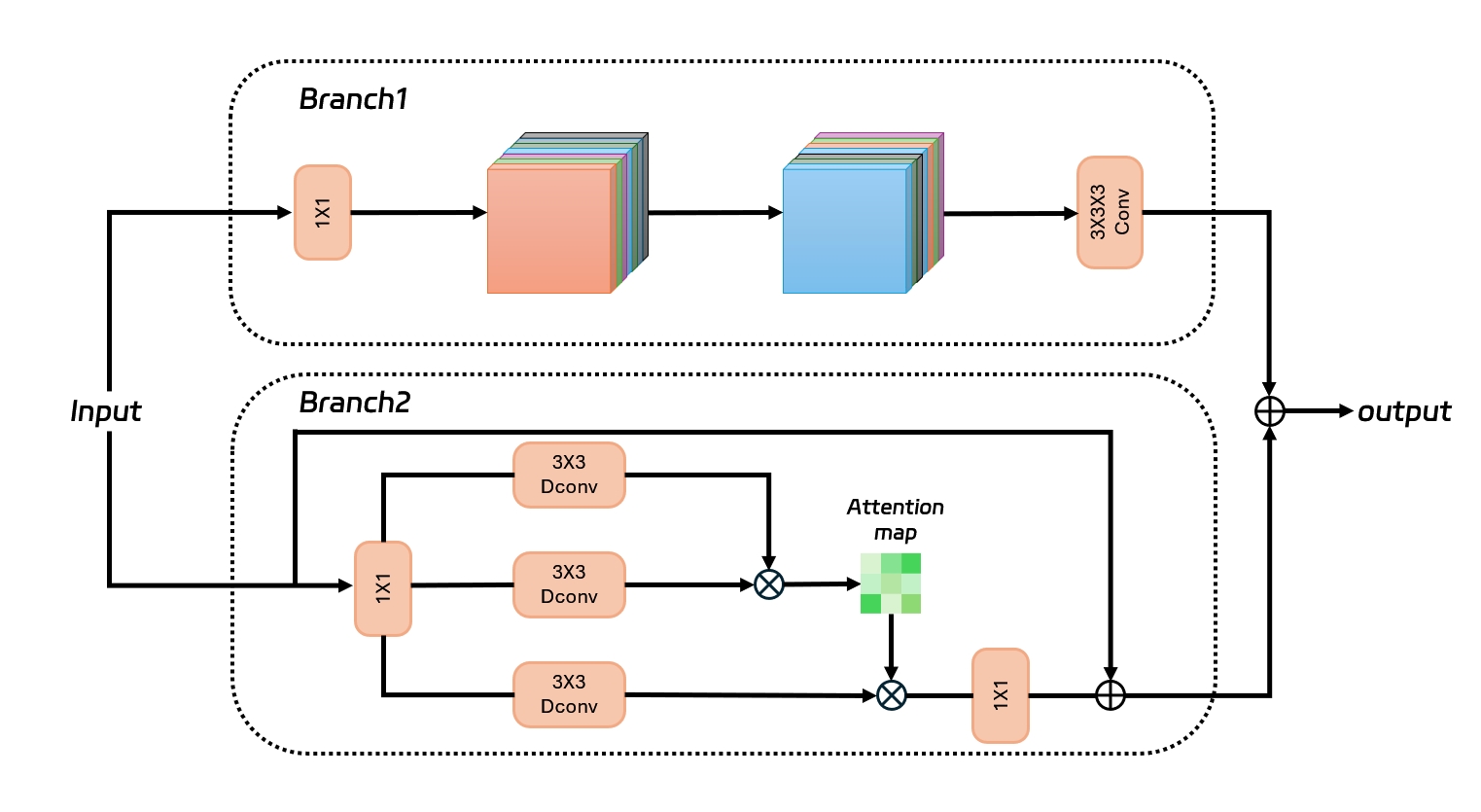}
	\caption{Structure of the proposed DAFM dual-attention fusion module}
	\label{fig:evaluation}
\end{figure}

Convolutional operations are inherently local and limited in receptive field, making it difficult to model global dependencies. In contrast, Transformer-style attention mechanisms are effective at capturing global context and long-range interactions. To jointly leverage local and global information, we design the Dual-Attention Fusion Module (DAFM), whose structure is shown in Fig. 2. DAFM uses two complementary branches to capture local details and global context, and fuses them to generate refined feature maps. Branch 1 focuses on learning noise-suppressed local features, whereas Branch 2 applies self-attention to capture long-range dependencies.

In Branch 1, the input feature map \(Y\) is first passed through a \(1 \times 1\) convolution to adjust the channel dimension. A channel shuffle (CS) operation is then applied: the channels are divided into groups, depthwise separable convolutions are performed within each group, and the outputs are concatenated along the channel dimension. A \(3 \times 3\) convolution is finally used to extract local features. The overall transform in Branch 1 can be written as
$$
F_{\text{conv}} = W_{3 \times 3}\bigl(\mathrm{CS}(W_{1 \times 1}(Y))\bigr),
$$
where \(F_{\text{conv}}\) denotes the output of Branch 1, \(W_{1 \times 1}\) and \(W_{3 \times 3}\) denote \(1 \times 1\) and \(3 \times 3\) convolution operators, respectively, and \(\mathrm{CS}(\cdot)\) denotes the channel shuffle operation.

In Branch 2, a \(1 \times 1\) convolution followed by a \(3 \times 3\) depthwise convolution generates three feature maps \(Q\), \(K\) and \(V\), each of size \(H \times W \times C\). To reduce computational cost, two of the tensors are reshaped to \((H W) \times C\) and \(C \times (H W)\), respectively, to compute a channel-wise attention map of size \((H W) \times (H W)\), which captures the interaction across spatial locations. This attention map reflects the relative importance between different feature positions and channels; for example, some channels may respond mainly to anthracnose lesions, whereas others are more sensitive to tea sooty mold. By learning the relationships between channels, the model can enhance responses in lesion-related channels and suppress background-dominated channels. The output of Branch 2 is
$$
F_{\text{att}} = W_{3 \times 3} \cdot \mathrm{Attention}(Q, K, V) + Y,
$$
where
$$
\mathrm{Attention}(Q, K, V) = V \cdot \mathrm{Softmax}\left(\frac{K Q}{a}\right),
$$
and \(a\) is a learnable scaling parameter that controls the magnitude of the \(KQ\) product before the Softmax operation.

Finally, the local detail features from Branch 1 and the global contextual features from Branch 2 are fused by element-wise addition to form the output of DAFM:
$$
F_{\text{out}} = F_{\text{conv}} + F_{\text{att}}.
$$
By combining local convolutions and global attention in a dual-branch structure, DAFM can precisely describe lesion edges and textures while capturing global context, and it allocates computation to regions that are most informative for tea pest and disease recognition. Using channel-wise attention instead of full spatial attention also reduces computational overhead, enabling performance gains without sacrificing the real-time detection speed of YOLOv8.

\subsection{Detect-OAHead: Occlusion-aware detection head}
Occlusion is a central challenge in tea leaf pest and disease recognition. When lesions or insects are partially occluded by other leaves and branches, their visual patterns are incomplete. A standard convolutional detector may fail to extract sufficient discriminative features, leading to erroneous predictions. In the original YOLOv8 architecture, the detection head consists of several standard convolutional layers that directly perform classification and bounding-box regression on the multi-scale feature maps output by the neck. These layers lack the ability to dynamically focus on informative regions and explicitly reason about occlusion relationships.

To overcome this limitation, we enhance the YOLOv8 detection head to make it occlusion-aware. The core idea of the Occlusion-Aware Head (OAHead), whose structure is shown in Fig. 3, is to learn the relationship between visible and occluded parts of the targets and to compensate for the feature loss caused by occlusion. Compared with the original head, we embed OAHead into both the classification and regression branches. The first part of OAHead is a depthwise separable convolution with residual connections. Depthwise separable convolution performs spatial convolution independently on each channel, allowing the model to learn spatial correlations at multiple receptive field sizes while keeping the parameter count low. However, depthwise separable convolution alone may overlook dependencies across channels. To address this, the outputs of the depthwise convolutions are fused via a pointwise \(1 \times 1\) convolution, and a two-layer fully connected network is used to model interactions among channels, enabling the network to enhance channel-wise correlations. Leveraging the learned relationship between occluded and unoccluded lesion features, OAHead produces a set of attention weights that highlight informative features and suppress irrelevant ones. The OAHead output is then multiplied with the original features as a gating mechanism, so that the model can better handle occluded lesions.

By replacing the original convolutional layers in the detection head with OAHead, our Detect-OAHead head performs a "pre-processing" step on the feature maps prior to final classification and regression. As a result, both localization and classification decisions are based on reweighted feature maps with more complete information. Equipped with OAHead, Detect-OAHead is no longer a passive decoder of features, but an active analyzer that is robust to occlusion, which improves the robustness and accuracy of DAONet-YOLOv8 in dense and cluttered tea plantations.

\begin{figure}[htbp]
	\centering
	\includegraphics[width=\columnwidth]{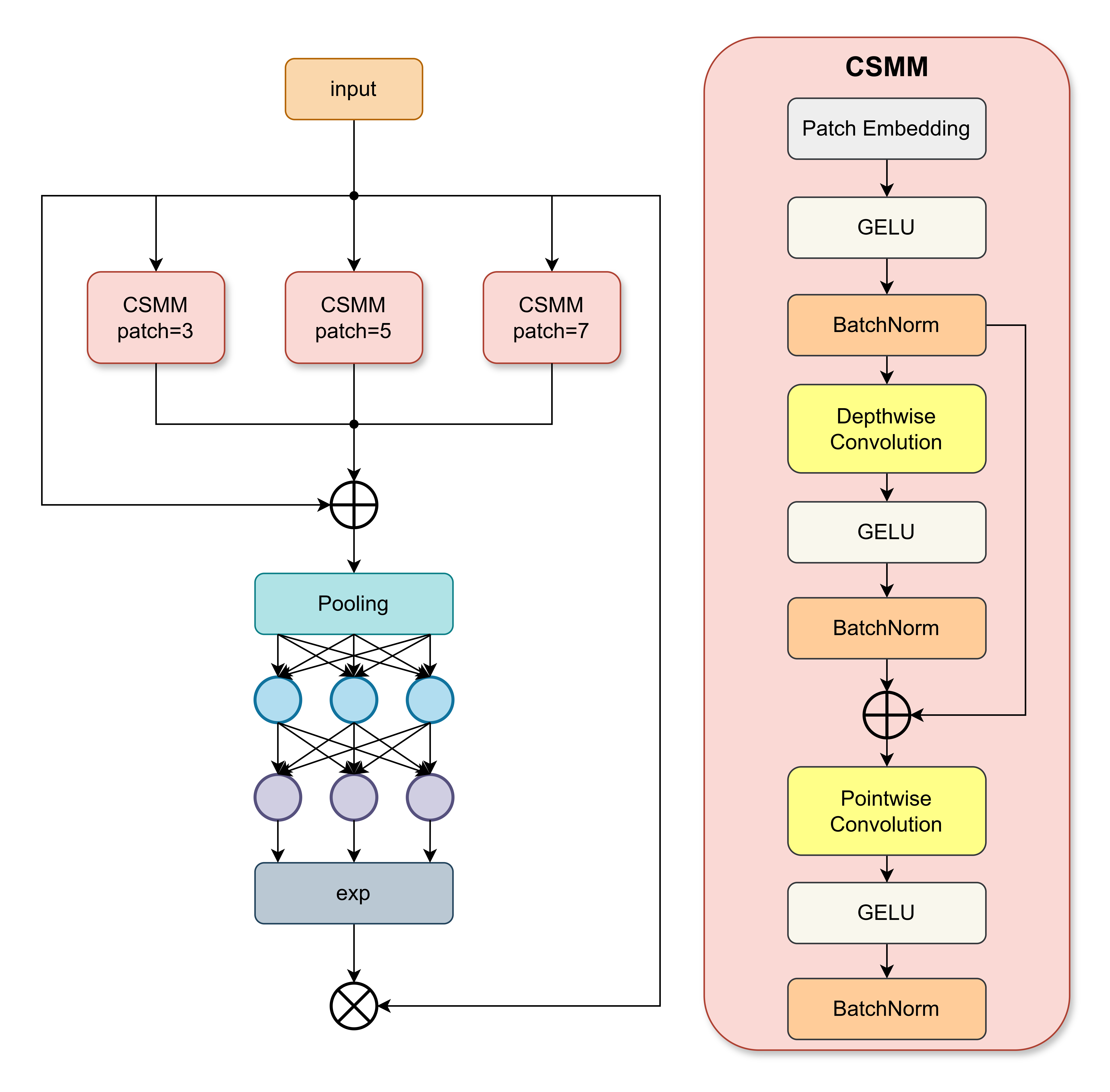}
	\caption{Structure of the OAHead occlusion-aware module}
	\label{fig:evaluation}
\end{figure}

\subsection{C2f-DSConv: Dynamic synthesis convolution module}
To enhance the representation ability of the model for diverse lesion shapes of tea leaf pests and diseases, we design the C2f-DSConv module. The core idea is to replace the standard bottleneck in the original C2f block with a Dynamic Synthesis Convolution (DSConv) module, as shown in Fig. 4. C2f-DSConv preserves the cross-stage local feature fusion advantages of C2f while significantly improving the flexibility and adaptivity of feature extraction.

DSConv is not a single fixed convolution, but a composite structure consisting of multiple parallel depthwise separable convolution branches and a dynamic weight generation mechanism. The branches employ kernels of different shapes, such as square k x k kernels and vertical and horizontal strip kernels of sizes m x 1 and 1 x m. The square kernels focus on local details and generic edges, whereas the strip kernels are more suitable for capturing elongated structures and linear features along vertical or horizontal directions. The branches operate in parallel on the same input and each produces a feature response, denoted as \(F_{\text{square}}\), \(F_{\text{vertical}}\) and \(F_{\text{horizontal}}\), respectively.

After obtaining these branch outputs, DSConv performs global average pooling on the feature maps to obtain channel-wise descriptors that summarize the global distribution of the input. These descriptors are fed into a fully connected layer to produce a score vector along the channel dimension, which is then normalized to obtain the importance weights \(\alpha_1\), \(\alpha_2\) and \(\alpha_3\) for the square, vertical and horizontal branches. The weights vary with each input during forward propagation and determine the relative contribution of each branch, forming a dynamic kernel synthesis mechanism driven by global feedback. The final synthesized feature map \(F_{\text{synthesis}}\) is computed as
$$
F_{\text{synthesis}} = \alpha_1 \cdot F_{\text{square}} + \alpha_2 \cdot F_{\text{vertical}} + \alpha_3 \cdot F_{\text{horizontal}}.
$$
This process is equivalent to dynamically constructing an optimal anisotropic composite convolution kernel for each input. The model is no longer limited to a fixed receptive field shape; instead, it can adjust its focus according to irregular lesion boundaries and varying insect poses, enabling more precise feature extraction for diverse targets. Thanks to its input-adaptive nature, DSConv also improves robustness to rotations, pose changes and mild deformations: regardless of the orientation of the lesions or insects, the dynamic synthesis mechanism can find an appropriate combination of kernels.
\begin{figure}[htbp]
	\centering
	\includegraphics[width=\columnwidth]{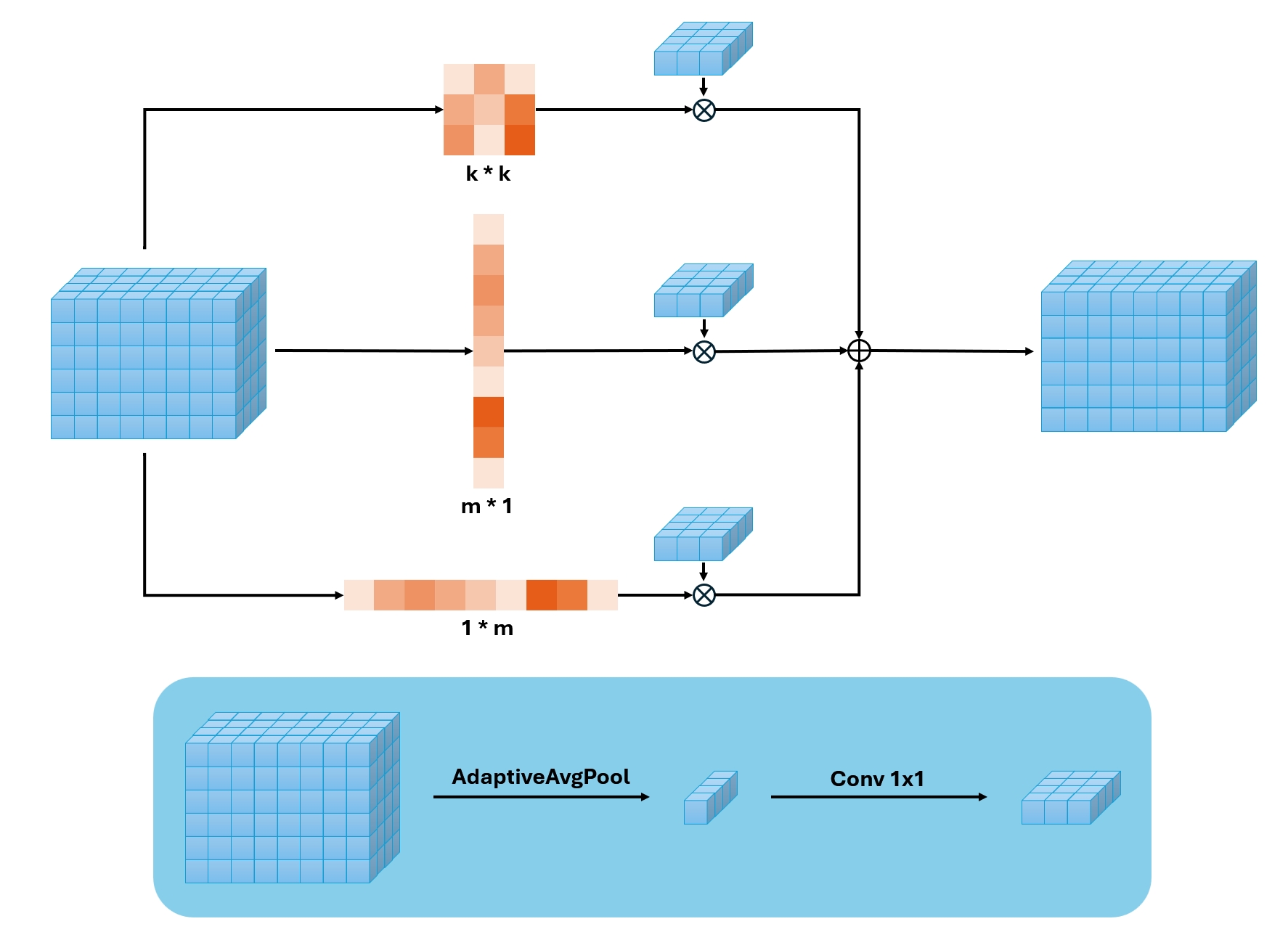}
	\caption{Structure of the proposed C2f-DSConv dynamic synthesis convolution module}
	\label{fig:evaluation}
\end{figure}

\section{Experiments Setup}

\subsection{Dataset}
Currently, only a few datasets are available for tea leaf pest and disease detection. Public datasets such as teaLeafBD: Tea Leaf Disease Detection\cite{alam2025tealeafbd} and Identifying Disease in Tea Leaves suffer from a clear limitation: most images are collected under controlled indoor lighting with single-leaf samples, and each image focuses on a single tea leaf. Although this acquisition setup highlights disease patterns, it fails to faithfully reflect the complex background illumination, occlusions among branches and leaves, and the spatial distribution of lesions over entire tea plants in real plantations. Constructing a realistic and usable dataset for tea leaf pests and diseases is therefore particularly important.

In this work, we collect images in July 2025 from a tea plantation in Youfangdian, Jinzhai County, Anhui Province, China. The weather was sunny with moderate sunlight, which is representative of typical field conditions. We use a Canon EOS 80 digital SLR camera with an image resolution of 6000x4000 pixels. The dataset covers six categories: anthracnose, red leaf disease, tea sooty mold, tea geometrid pest, white spot disease and healthy leaves, with a total of 2496 images. Because the original images are very high resolution and unsuitable for direct training, we crop and resize them before model training.

\subsection{Evaluation Metrics}
All experiments are conducted on a Windows 11 system using the PyTorch framework. The GPU is an NVIDIA GeForce RTX 4060, with Python 3.10 and CUDA 12.1 as the software environment.

To objectively and comprehensively evaluate the performance of the proposed model, we adopt precision, recall, mAP@50, mAP@50:95, the number of parameters (Params) and GFLOPs as evaluation metrics. For mAP@50, the IoU threshold is fixed at 0.5. For mAP@50:95, the IoU threshold is varied from 0.5 to 0.95 with a step size of 0.05, and the mean AP over these 10 IoU thresholds is reported.

\section{Experiments and Result}
\subsection{Ablation Experiments}

\begin{table*}[t]
\centering
\caption{Ablation study of DAONet-YOLOv8 on the tea leaf pest and disease dataset.}
\small
\begin{tabular}{ccccccccc}
\hline
DAFM & OAHead & DSConv & Precision & Recall & mAP@50 & mAP@50:95 & Params (M) & GFLOPs \\
\hline
 &  &  & 0.9063 & 0.8812 & 0.9570 & 0.7510 & 3.0 & 8.1 \\
\checkmark &  &  & 0.9171 & 0.9177 & 0.9670 & 0.7583 & 3.4 & 8.4 \\
 & \checkmark &  & 0.9074 & 0.9078 & 0.9590 & 0.7556 & 2.8 & 7.0 \\
 &  & \checkmark & 0.9159 & 0.9199 & 0.9683 & 0.7355 & 2.3 & 6.3 \\
\checkmark & \checkmark &  & 0.9193 & 0.9226 & 0.9657 & 0.7617 & 3.1 & 7.3 \\
\checkmark &  & \checkmark & 0.9170 & 0.9110 & 0.9694 & 0.7636 & 2.7 & 6.6 \\
 & \checkmark & \checkmark & 0.9168 & 0.9297 & 0.9604 & 0.7585 & 2.2 & 5.3 \\
\checkmark & \checkmark & \checkmark & 0.9297 & 0.9280 & 0.9710 & 0.7690 & 2.5 & 5.5 \\
\hline
\end{tabular}
\end{table*}

From Table 1 we can observe that adding the DAFM attention module noticeably improves precision. By combining convolutional local features with global context modeling via self-attention, the dual-branch structure allows the model to focus more on key lesion regions, albeit at the cost of slightly increased computation. Building on this, replacing the vanilla head with the Detect-OAHead occlusion-aware head further improves recall and mAP@50:95 by 4.49 and 0.34 percentage points, respectively. More importantly, while performance increases, the computational cost decreases from 8.4 GFLOPs to 7.3 GFLOPs and the number of parameters is reduced by about 3.1\%, demonstrating that Detect-OAHead can effectively learn the relationship between occluded and visible regions, compensate for occlusion-induced feature loss and simultaneously optimize computational efficiency. Finally, replacing the C2f blocks in the backbone with C2f-DSConv yields our final DAONet-YOLOv8 model. When the three modules work together, all metrics improve significantly. Compared with the YOLOv8n baseline, precision, recall, mAP@50 and mAP@50:95 increase by 2.34, 4.68, 1.40 and 1.80 percentage points, respectively. These results indicate that C2f-DSConv effectively enhances the recognition of irregular lesion shapes while reducing model complexity, and in combination with DAFM and Detect-OAHead, it enables high accuracy and robustness under complex backgrounds and occlusions.

\subsection{Comparative Experiments}

\begin{table*}[t]
\centering
\caption{Comparison of DAONet-YOLOv8 with other object detection models.}
\small
\begin{tabular}{ccccccc}
\hline
Model         & Precision & Recall & mAP@50 & mAP@50:95 & Params (M) & GFLOPs \\
\hline
Fast R-CNN\cite{ren2015faster}    & 0.9023    & 0.8356 & 0.9018 & 0.6785    & 35.7       & 180.5  \\
YOLOv5\cite{jocher2020yolov5}        & 0.8961    & 0.8695 & 0.9449 & 0.7211    & 2.5        & 7.1    \\
YOLOv9\cite{wang2024yolov9}         & 0.9049    & 0.8672 & 0.9585 & 0.7174    & 2.0        & 7.6    \\
YOLOv10\cite{wang2024yolov10}       & 0.8579    & 0.7702 & 0.9012 & 0.6700    & 2.3        & 6.5    \\
YOLOv11\cite{khanam2024yolov11}       & 0.9242    & 0.9232 & 0.9592 & 0.6967    & 2.6        & 6.3    \\
YOLOv12\cite{tian2025yolov12}       & 0.8678    & 0.8289 & 0.9289 & 0.7193    & 2.6        & 6.3    \\
DAONet-YOLOv8 & 0.9297    & 0.9280 & 0.9710 & 0.7690    & 2.5        & 5.5    \\
\hline
\end{tabular}
\end{table*}

From Table 2, Fast R-CNN as a representative two-stage detector performs significantly worse than YOLO-based one-stage models in both accuracy and efficiency, and its high computational cost makes real-time tea plantation monitoring impractical. This further confirms the advantage of one-stage detectors for agricultural pest and disease detection.

Within the YOLO family, DAONet-YOLOv8 also demonstrates superior overall performance. Compared with the widely used YOLOv5 model, DAONet-YOLOv8 improves precision, recall, mAP@50 and mAP@50:95 by 3.36, 5.85, 2.61 and 4.79 percentage points, respectively, while achieving a similar parameter count and reducing computational complexity by about 22.5\%. This shows that DAONet-YOLOv8 simultaneously improves accuracy and efficiency. When compared with recent YOLO variants on the tea pest and disease detection task, DAONet-YOLOv8 remains competitive. YOLOv10 has slightly fewer parameters but much lower precision and recall (0.8579 and 0.7702), which are insufficient for practical tea pest and disease monitoring. YOLOv11 is close to DAONet-YOLOv8 in terms of precision and recall, but its mAP@50:95 is only 0.6967, much lower than the 0.7690 achieved by DAONet-YOLOv8, indicating insufficient accuracy at high IoU thresholds. YOLOv12 also lags behind DAONet-YOLOv8 in precision, recall and mAP@50:95. Overall, DAONet-YOLOv8 achieves the best trade-off between accuracy and efficiency among all compared models.

\subsection{Visualization Analysis}
To qualitatively evaluate the behavior of the model in complex tea plantation scenes, we visualize detection results on representative samples. The selected image contains typical red leaf lesions, with leaves partially occluded by interlaced branches and leaves, and a bright, textured background that provides strong interference. We compare the predicted bounding boxes and Grad-CAM activation maps\cite{selvaraju2017grad} of the YOLOv8n baseline and the proposed DAONet-YOLOv8, as shown in Fig. 5.

\begin{figure}[htbp]
	\centering
	\includegraphics[width=\columnwidth]{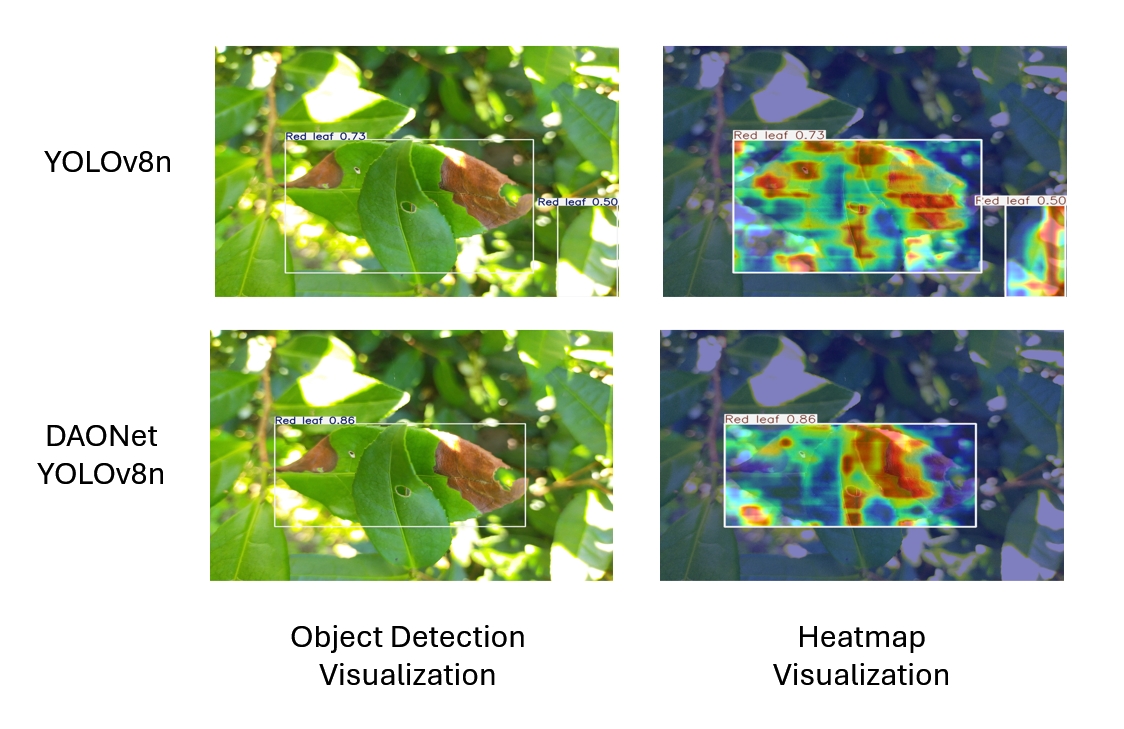}
	\caption{Visualization of detection results and Grad-CAM activation maps for YOLOv8n and DAONet-YOLOv8}
	\label{fig:evaluation}
\end{figure}

YOLOv8n produces two bounding boxes with confidence scores of 0.73 and 0.50, showing instability in lesion boundary delineation. In contrast, DAONet-YOLOv8 outputs a single tight bounding box with confidence 0.86, significantly reducing redundant detections. The Grad-CAM heatmaps reveal that YOLOv8n spreads attention across irrelevant background regions, while DAONet-YOLOv8 concentrates responses on lesion cores and edges, effectively suppressing background interference. These results validate that the proposed modules jointly enhance robustness to complex backgrounds and occlusions.

\section{Conclusion}
We proposes DAONet-YOLOv8, a novel object detection model tailored for tea leaf pest and disease detection in complex plantation environments. The model integrates three key components: a Dual-Attention Fusion Module (DAFM) that combines local convolutions with global self-attention to enhance sensitivity to subtle lesion features while suppressing background noise, an occlusion-aware detection head (Detect-OAHead) that learns visible-occluded relationships to compensate for feature loss caused by dense foliage, and a C2f-DSConv module with dynamic kernel synthesis to adaptively capture irregular lesion shapes. Experiments on our real-world tea plantation dataset covering six categories demonstrate that DAONet-YOLOv8 improves precision, recall, mAP@50 and mAP@50:95 by 2.34, 4.68, 1.40 and 1.80 percentage points over YOLOv8n, while reducing parameters by 16.7\%. Comparative experiments against Fast R-CNN and recent YOLO variants confirm that DAONet-YOLOv8 achieves the best accuracy efficiency trade-off, making it suitable for real-time deployment in intelligent tea plantation monitoring systems.


\bibliographystyle{IEEEtran}
\bibliography{references}

\end{document}